\pdfoutput=1

\documentclass[11pt]{article}

\usepackage[]{ACL2023}

\usepackage{times}
\usepackage{latexsym}

\usepackage[T1]{fontenc}

\usepackage[utf8]{inputenc}

\usepackage{microtype}

\usepackage{inconsolata}
\usepackage{multirow}
\usepackage{amssymb,amsfonts,amsmath}
\usepackage{upgreek}
\usepackage{graphicx}
\usepackage{bm}
\usepackage{listings}
\usepackage{needspace}
\title{GKD: A General Knowledge Distillation Framework for Large-scale Pre-trained Language Model}

\author{
  Shicheng Tan${}^{*1}$,~
  Weng Lam Tam${}^{2}$,~
  Yuanchun Wang${}^{3}$,~
  Wenwen Gong${}^{4}$,
  \\
  {\bf Shu Zhao\footnotemark[2] ${}^{1}$},~
  {\bf Peng Zhang${}^{2}$},~
  {\bf Jie Tang\footnotemark[2] ${}^{4}$}
  \\
  ${}^{1}$Anhui University,~
  ${}^{2}$Zhipu.AI,~
  ${}^{3}$Renmin University of China,~
  ${}^{4}$Tsinghua University \\
  {\tt tsctan@foxmail.com,\{rainatam9784,frederickwang99\}@gmail.com} \\
  {\tt wenweng@mail.tsinghua.edu.cn,zhaoshuzs2002@hotmail.com} \\
  {\tt peng.zhang@zhipuai.cn,jietang@tsinghua.edu.cn} \\
}

\begin{document}
\maketitle
\renewcommand{\thefootnote}{\fnsymbol{footnote}}
\footnotetext[1]{This work was done when the author visited Zhipu.AI.}
\footnotetext[2]{Corresponding authors.}
\footnotetext[0]{The other authors also include Yang Yang, Hongyin Tang, Keqing He, Jiahao Liu, and Jingang Wang from Meituan.}
\renewcommand{\thefootnote}{\arabic{footnote}}

\begin{abstract}
Currently, the reduction in the parameter scale of large-scale pre-trained language models (PLMs) through knowledge distillation has greatly facilitated their widespread deployment on various devices. However, the deployment of knowledge distillation systems faces great challenges in real-world industrial-strength applications, which require the use of complex distillation methods on even larger-scale PLMs (over 10B), limited by memory on GPUs and the switching of methods. To overcome these challenges, we propose GKD, a general knowledge distillation framework that supports distillation on larger-scale PLMs using various distillation methods. With GKD, developers can build larger distillation models on memory-limited GPUs and easily switch and combine different distillation methods within a single framework. Experimental results show that GKD can support the distillation of at least 100B-scale PLMs and 25 mainstream methods on 8 NVIDIA A100 (40GB) GPUs. 
\footnote{The code is available at \url{https://github.com/aitsc/GLMKD}.}
\end{abstract}

\section{Introduction}
Pre-trained language models, such as BERT \citep{RN289}, RoBERTa \citep{RN1181}, and their variants, have achieved excellent success in natural language processing (NLP) tasks when they usually have hundreds of millions of parameters. Considering computationally expensive resource constraints, a wide range of real-world applications are often impeded. Knowledge distillation \citep{RN4609}, as a method for compressing large-scale pre-trained language models, is attracting more and more attention. As large-scale PLMs continue to grow in scale, and with advancements in knowledge distillation methods, it becomes increasingly pressing to apply knowledge distillation research in controlled laboratory settings to the real world.

The field of knowledge distillation for language models has witnessed a phenomenal progress in recent years, particularly with regards to the reduction of model size, leading to the development of a plethora of sophisticated distillation techniques \citep{RN7341,RN7325} and a comprehensive toolkit \citep{RN7468}. However, despite these rich research outcomes, there are still major challenges in deploying knowledge distillation systems for real-world industrial-strength applications, including:

\begin{itemize}
\item \textbf{Obstacles to Distilling Ultra-large-scale PLMs.} Contrary to distillation in controlled laboratory settings aimed at models with billions of parameters, many industrial-strength applications \citep{RN7516} rely on ultra-large-scale PLMs (on the order of 10B or even larger). The training of ultra-large-scale PLMs is already challenging, and the distillation process requires simultaneous training of both large and small models, leading directly to difficulties in distillation of ultra-large-scale PLMs. Furthermore, there are also methods \citep{RN7379,RN7370} for distilling multiple large models into a single small model, which pose significant challenges in memory-constrained GPU environments.
\item \textbf{Obstacles to Switching Distillation Methods.} Deploying a knowledge distillation system requires the implementation of numerous distillation methods to meet different requirements, but due to the differences in implementation of these methods, it is difficult to switch and combine them easily within a framework. It is important to have an architecture that accommodates a range of distillation methods while ensuring efficient training, such as avoiding excessive extraction of intermediate features that lead to memory waste. Thus, a compatible and efficient architecture is crucial for successful deployment of knowledge distillation systems.
\end{itemize}

To overcome these challenges, we present a general knowledge distillation framework (GKD) for deploying knowledge distillation systems that support various scale PLMs and methods. To overcome the obstacles to distilling ultra-large-scale PLMs, GKD leverages the techniques of training large transformer models to the distillation process that requires training multiple large (teacher) and small (student) models simultaneously, incorporating the latest model and data parallel strategies. To overcome the obstacles to switching distillation methods, GKD employs a dynamic hook mechanism and auxiliary model to extract and operate intermediate layer features and inference process of models in each iteration. While being compatible with various methods, it avoids the waste of memory caused by extracting all intermediate layers. GKD presents the first exploration of knowledge distillation for language models in industrial scenarios. Specifically, our main contribution lies in:

\begin{itemize}
\item \textbf{Larger-scale Model Distillation.} We propose a teacher-student parallel strategy based on advanced memory optimization methods, addressing the challenge of distilling ultra-large-scale PLMs (over 10B) due to memory constraints. The proposed strategy supports distillation of at least 100B-scale PLMs on 8 NVIDIA A100 (40GB) GPUs.
\item \textbf{More Compatible Method Architecture.} We propose an efficient adaptive architecture compatible with various methods, addressing the challenge of switching and using different distillation methods within a single framework with difficulty. The proposed architecture supports at least 25 model distillation methods.
\item \textbf{Easy-to-use Open Source Toolkit.} We have open-sourced the required toolkit for GKD, which provides a command-line interface for 25 distillation methods, facilitating developers to deploy knowledge distillation systems for ultra-large-scale PLMs. 
\end{itemize}

\section{Related work}
In recent years, knowledge distillation for compressing PLMs has gained increased attention. These works studied ways of better utilizing language model features for transferring knowledge from large teacher models to a smaller student model, involving hidden layers \citep{RN7482}, attention layers \citep{RN7385}, soft labels \citep{RN5665}, and hard labels \citep{Continuation}.
These works validated their methods with PLMs of hundreds of millions of parameters, such as BERT \citep{RN289}, RoBERTa \citep{RN1181}, XLNet \citep{XLNet}, etc. However, deployment of the distillation system on GPUs with limited memory has been hindered by the reliance on ultra-large-scale PLMs (10B or even larger). An offline distillation method \citep{RN7414} that saved teacher features before training the student individually reduced memory pressure, but was limited to methods with smaller feature scales and without teacher-student interaction. In this work, GKD was compatible with ultra-large-scale PLMs distillation via the introduction of Megatron-LM \citep{RN2180} based on model parallelism and Zero Redundancy Optimizer (ZeRO) \citep{RN2184} based on data parallelism.

While some code for knowledge distillation methods focused on language models was made public \citep{RN6418,RN7482,RN7111}, there was a lack of a general framework for deploying knowledge distillation systems. TextBrewer \citep{RN7468} packaged some abstract and simple distillation processes and loss functions, but lacked implementation of many methods and was difficult to adapt to increasingly complex distillation methods. There were significant differences in the implementation of these methods, such as DIITO \citep{RN7325} requiring dynamic intervention of the intermediate layer computation in the model; SID \citep{RN7487} changing the intermediate layer features during training; Continuation-KD \citep{Continuation} altering the loss calculation method as the epoch increased, and so on. These differences in implementation made it difficult for them to be easily switched and combined within a single framework, hindering the application of various advanced methods in knowledge distillation systems. In this work, GKD accommodated various advanced knowledge distillation methods through a dynamic hook mechanism and auxiliary models.

\begin{figure*}
  \centering
  \includegraphics[width=\linewidth,scale=1]{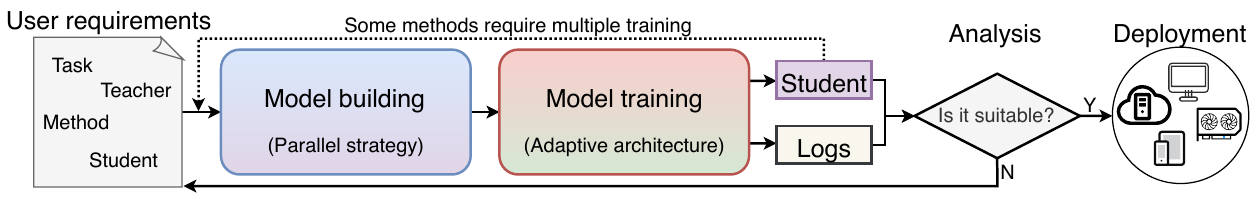}
  \caption{The framework of the GKD. From the user requirements to the model deployment on the device, the GKD includes the six main processes involved in the deployment of the knowledge distillation system. \label{fig:framework}}
\end{figure*}

\section{GKD}
In this section, we first introduce the overview framework of the proposed GKD, then delve into the details of how GKD implements larger-scale model distillation and a more compatible method architecture, from the perspective of model building and training.

\subsection{Overview Framework}
Figure~\ref{fig:framework} shows the overview framework of GKD, which consists of six main processes:

(1) \emph{User Requirements:} This process begins with the user specifying their requirements and forming a configuration file, which includes the choice of training task, distillation method, teacher model, student model, etc.

(2) \emph{Model Building:} This process addresses the obstacles to distilling ultra-large-scale PLMs by implementing a teacher-student parallel strategy that combines Megatron-LM \citep{RN2180} and ZeRO \citep{RN2184}. The process involves selecting and executing parameter initialization strategies for the student model, such as initializing the student model with a pre-trained student, a truncated parameter teacher, random initialization methods, or other distilled students. It also includes initializing the training data with a tokenizer.

(3) \emph{Model Training:} This process addresses the obstacles to switching distillation methods by implementing an efficient adaptive architecture that is compatible with various methods. This process includes the initialization of methods to extract and compute different model features based on the requirements of different methods at different iteration numbers. 

(4) \emph{Multiple Training:} This process is utilized for methods that require multiple training, such as task-specific methods \citep{RN7482} that necessitate distillation in the task-specific stage after distillation in the pre-training stage.

(5) \emph{Analysis:} This process confirms the compliance of the distilled student model with deployment requirements through analysis, such as examining the performance on the test set and other phenomena that can be utilized to enhance the model.

(6) \emph{Deployment:} This process deploys the student model on the corresponding device, such as low-computing mobile devices or services with higher load deployment under equal computing power.

These six processes are performed in sequence to form the workflow of the knowledge distillation system. The greatest contribution of GKD lies in the design of the model building and training, as the other processes do not pose a challenge to the deployment of the knowledge distillation system. In the following sections, we will provide a detailed description of how GKD enables larger-scale model distillation in the model building and more compatible method architectures in the model training.

\subsection{Model Building}
The challenge in building models lies in allocating ultra-large-scale PLMs, consisting of a student and one or more teacher models, on a GPU with only several tens of GB of memory. To address this challenge, we propose a teacher-student parallel strategy that splits the model parameters to different GPUs while preserving the feature distance computation between the teacher and student models. This strategy is inspired by the optimization of single ultra-large-scale PLMs, including Megatron-LM \citep{RN2180} which splits each parameter matrix in the transformer across multiple GPUs, and ZeRO \citep{RN2184} which partitions each layer of transformers sequentially across multiple GPUs.

\begin{figure}
  \centering
  \includegraphics[width=\linewidth,scale=1]{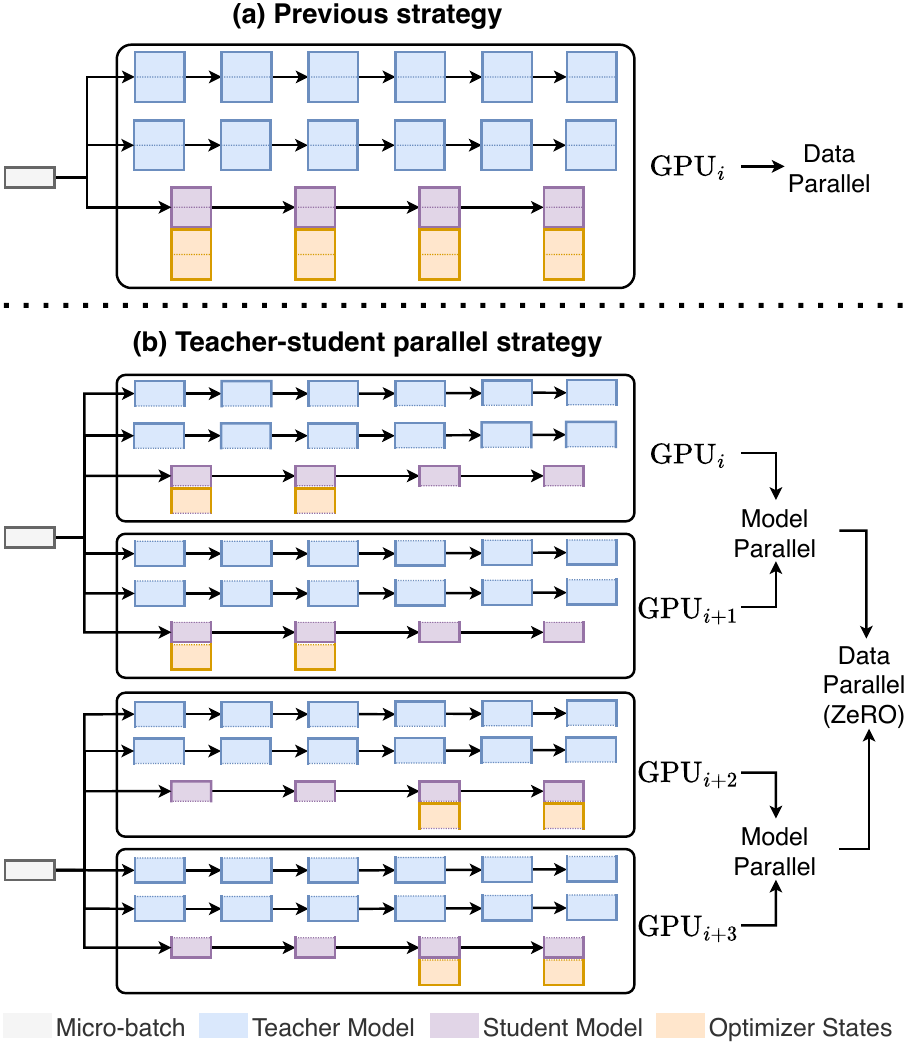}
  \caption{This comparison between the previous strategy and the proposed teacher-student parallel strategy is demonstrated through an example, where it can be observed that the teacher-student parallel strategy significantly reduces the memory utilization of each GPU. \label{fig:parallel}}
\end{figure}

As shown in Figure~\ref{fig:parallel}, we demonstrate the comparison between the previous strategy and our proposed teacher-student parallel strategy using an example. The example includes the allocation of two 6-layer transformer teacher models and one 4-layer transformer student model on the GPU. The current methods allocate all the model parameters on each GPU, severely limiting the training of ultra-large-scale PLMs and multiple models. To reduce the memory usage on each GPU without compromising the interaction between the teacher and the student, our teacher-student parallel strategy evenly distributes the parameters of the teacher and student on different GPUs, with each GPU corresponding to the matching parameters of the teacher and student. With the model parallel and data parallel count being 2, the memory usage can be reduced by at least half. If utilizing ZeRO-Offload \citep{ZeRO-Offload}, the optimizer states can further be stored in CPU memory to reduce the utilization of GPU memory.

\subsection{Model Training}
\begin{figure}
  \centering
  \includegraphics[width=0.7\linewidth,scale=1]{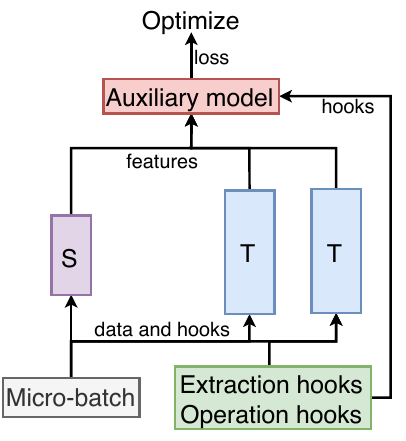}
  \caption{A workflow for efficient adaptive architecture compatible with various methods in a single iteration. \label{fig:architecture}}
\end{figure}

The challenge in training models lies in how to easily switch and use different distillation methods within a single framework. To address this challenge, we propose an efficient adaptive architecture that is compatible with various methods. It implements the operation of different methods and the calculation of features through a dynamic hook mechanism and an auxiliary model, respectively. As shown in the workflow in Figure~\ref{fig:architecture}, the dynamic hook mechanism constructs extraction hooks for extracting model features and operation hooks for modifying the model inference process during each iteration. These hooks are described by a configuration file similar to JSON, which only requires recording the operations required by the method and playing a role during the model inference process. The auxiliary model calculates the loss function based on these hooks and the returned model features. Table~\ref{tab:architecture} describes the features that this architecture can adapt to existing methods.

It is worth noting that GKD can achieve method combination by integrating hooks from different methods.
GKD can also record all model features through extraction hooks and save the distance of teacher and student features in the auxiliary model for later analysis of the correlation between feature distance and task performance in the distillation process.

\begin{table*}
\centering
\resizebox{\linewidth}{!}{
\begin{tabular}{p{6cm}p{14cm}} \hline
    \textbf{Compatible features} & \textbf{Representative methods} \\ \hline
     Modify the inference process of the model & DIITO \citep{RN7325}, LRC-BERT \citep{RN7437}, Theseus \citep{RN7113} \\ 
     Dynamically modify the feature extraction or inference process & SID \citep{RN7487}, Theseus \citep{RN7113} \\ 
     Additional trainable parameters & TinyBERT \citep{RN7482}, RAIL-KD \citep{RN7350}, Universal-KD \citep{RN7376}, LRC-BERT \citep{RN7437} \\ 
     Dynamically change loss function & Annealing-KD \citep{RN5665}, Continuation-KD \citep{Continuation}, MobileBERT \citep{RN7111} \\
     Complex intermediate layer calculation & CKD \citep{RN7407}, MGSKD \citep{RN7341}, ALP-KD \citep{RN7406} \\
     Train student by multiple teachers & TMKD \citep{RN6109}, MT-BERT \citep{RN7379}, RL-KD \citep{RN7370}, Uncertainty \citep{RN7419} \\
     Multiple training reduces teacher until student scale & TAKD \citep{RN7475}, DGKD \citep{RN7395} \\
     Other simple methods & KD \citep{RN4609}, PD \citep{RN6581}, PKD \citep{RN7494}, DistilBERT \citep{RN6418}, MiniLM \citep{RN7471}, MiniLMv2 \citep{RN7385} \\
    \hline
\end{tabular}
}
\caption{The compatible features and representative methods of our proposed adaptive architecture.}
\label{tab:architecture}
\end{table*}

\section{Experiments}
In this section, we verified that GKD, which is used for distillation of language models, can support at least 100B-scale parameters and 25 mainstream methods on 8 NVIDIA A100 (40GB) GPUs.

\subsection{Experimental Setup}
\paragraph{Datasets}
All methods that require distillation in the pre-training stage use BooksCorpus \citep{books} and English Wikipedia as training data (19GB). For the task-specific stage (fine-tuning), we evaluate different distillation methods using the more challenging SuperGLUE benchmark \citep{RN2189}.

\paragraph{Methods}
We tested 22 distillation methods specifically designed for language models, as well as three classic methods (KD, TAKD, and DGKD) from computer vision, which are listed in Tables \ref{tab:architecture} and \ref{tab:results}. The implementation of the teacher-student parallel strategy was carried out using the Megatron-LM \citep{RN2180} and DeepSpeed \citep{RN2185} framework.

\paragraph{Models}
The commonly used BERT \citep{RN289} lacks open-source ultra-large-scale PLMs, so we employed a more advanced GLM \citep{RN2172}, which boasts open-source models of 10B-scale or even 130B-scale \citep{glm130b}, significantly reducing the deployment cost of the knowledge distillation system. The scale of teachers and students are presented in Tables \ref{tab:results} and \ref{tab:memory-stat}.

Refer to Appendix \ref{sec:appendix:details} for more implementation details.

\subsection{Results}
\paragraph{More Compatible Method Architecture}
To verify the proposed adaptive architecture can effectively be compatible with various methods, we tested 25 mainstream distillation methods and present the results in Table \ref{tab:results}. The results demonstrate that these methods can be easily switched and utilized in GKD. It is worth noting that TinyBERT (without data augmentation) outperformed all the latest methods in our setup. This suggests that the latest methods may not necessarily be the most effective, and different requirements may necessitate different methods. Additionally, the reliability of GKD is further validated from the perspective of loss function values in Appendix \ref{sec:appendix:loss}.

\begin{table*}
\centering
\resizebox{\linewidth}{!}{
\begin{tabular}{l|ccccccccc} \hline
    \multirow{2}{*}{\textbf{Methods}} & \textbf{ReCoRD} & \textbf{COPA} & \textbf{WSC} & \textbf{RTE} & \textbf{BoolQ} & \textbf{WiC} & \textbf{CB} & \textbf{MultiRC} & \multirow{2}{*}{\textbf{avg}} \\ \cline{2-9}
     & \textbf{F1/Acc.}& \textbf{Acc.}& \textbf{Acc.}& \textbf{Acc.}& \textbf{Acc.}& \textbf{Acc.}& \textbf{F1/Acc.}& \textbf{F1$_a$/EM} & \\ \hline
    GLM$_\mathrm{Base}$ (teacher, 110M) & 72.80/72.17 & 66.00 & 77.88 & 72.92 & 79.39 & 66.14 & 88.19/91.07 & 72.32/26.34 & 71.72 \\
    GLM$_\mathrm{Large}$ (teacher, 340M) & 80.08/79.54 & 78.00 & 81.73 & 79.78 & 82.63 & 70.06 & 86.33/89.29 & 76.39/37.67 & 77.11 \\ \hline
    \multicolumn{10}{c}{\textbf{\textit{Single-teacher}}: Teacher (GLM$_\mathrm{Base}$) $\Rightarrow$ Student (66M)} \\ \hline
    KD \citep{RN4609} & 22.66/21.99 & 61.67 & 63.46 & 54.63 & 66.07 & 57.05 & 61.75/72.02 & 51.98/2.41 & 52.41 \\ 
    PD \citep{RN6581} & 54.36/53.59 & 65.67 & 66.67 & 59.45 & 69.82 & 59.20 & 80.13/81.55 & 65.97/15.29 & 62.03 \\ 
    PKD \citep{RN7494} & 61.77/60.99 & 60.00 & 65.38 & 68.83 & 77.73 & 65.78 & 82.76/85.12 & 69.99/22.67 & 66.17 \\ 
    DistilBERT \citep{RN6418} & 59.79/59.05 & 65.00 & 68.59 & 60.89 & 73.39 & 60.34 & 77.48/83.33 & 66.98/17.38 & 63.78 \\ 
    Theseus \citep{RN7113} & 57.07/56.33 & 61.67 & 66.35 & 68.11 & 77.81 & 64.37 & 89.14/87.50 & 69.08/21.79 & 66.09 \\ 
    TinyBERT \citep{RN7482} & 65.60/64.88 & 70.33 & \bf 75.00 & \bf 71.96 & \bf 77.97 & \bf 67.87 & \bf 89.58/89.88 & \bf 71.37/25.74 & \bf 70.83 \\ 
    MobileBERT$^\dagger$ \citep{RN7111} & 59.29/58.61 & 65.33 & 68.59 & 58.97 & 74.61 & 63.85 & 86.65/88.69 & 66.87/19.41 & 65.14 \\
    SID \citep{RN7487} & 27.17/26.19 & 65.00 & 65.06 & 58.12 & 69.33 & 57.16 & 51.02/73.81 & 59.26/14.55 & 55.08 \\ 
    MiniLM \citep{RN7471} & 60.00/59.24 & 62.00 & 63.46 & 67.63 & 75.88 & 64.99 & 67.63/79.17 & 67.36/19.66 & 63.81 \\ 
    MiniLMv2 \citep{RN7385} & 60.88/60.16 & 62.00 & 62.82 & 66.67 & 76.73 & 63.69 & 66.38/76.79 & 68.68/21.65 & 63.65 \\ 
    ALP-KD \citep{RN7406} & 57.72/56.90 & 60.67 & 64.74 & 68.11 & 77.20 & 64.79 & 74.82/79.76 & 68.21/19.90 & 64.27 \\ 
    LRC-BERT \citep{RN7437} & 55.10/54.44 & 65.67 & 66.67 & 56.56 & 74.86 & 57.63 & 80.27/81.55 & 65.75/16.16 & 62.25 \\ 
    Annealing-KD \citep{RN5665} & 56.08/55.39 & 69.33 & 66.67 & 58.97 & 70.57 & 59.82 & 85.78/85.12 & 66.26/13.92 & 63.33 \\ 
    CKD \citep{RN7407} & 56.35/55.65 & 65.00 & 66.67 & 61.25 & 71.63 & 58.83 & 88.61/84.52 & 66.11/15.22 & 63.33 \\ 
    Universal-KD \citep{RN7376} & 58.67/57.83 & 58.67 & 66.67 & 70.16 & 77.56 & 65.52 & 87.52/85.71 & 69.96/22.63 & 66.22 \\ 
    DIITO \citep{RN7325} & 63.71/63.00 & \bf 72.00 & 69.23 & 65.46 & 75.46 & 60.76 & 86.75/85.12 & 66.28/17.63 & 66.77 \\ 
    Continuation-KD \citep{Continuation} & 55.61/54.91 & 68.67 & 64.74 & 58.72 & 71.42 & 58.25 & 85.61/83.93 & 66.64/13.33 & 62.73 \\ 
    RAIL-KD \citep{RN7350} & 59.85/59.19 & 66.67 & 70.19 & 60.53 & 69.00 & 60.34 & 78.98/83.33 & 66.55/15.60 & 63.56 \\ 
    MGSKD \citep{RN7341} & 50.29/49.49 & 65.00 & 65.06 & 65.94 & 73.31 & 63.17 & 83.89/84.52 & 67.32/15.56 & 63.50 \\  \hline
    \multicolumn{10}{c}{\textbf{\textit{Multi-teacher}}: Teachers (GLM$_\mathrm{Base}$ and GLM$_\mathrm{Large}$) $\Rightarrow$ Student (66M)} \\ \hline
    TMKD \citep{RN6109} & \bf 65.77/65.09 & 70.33 & 63.14 & 66.91 & 75.37 & 63.38 & 70.22/79.17 & 68.76/22.77 & 65.63 \\ 
    MT-BERT \citep{RN7379} & 46.81/46.08 & 59.00 & 63.46 & 65.46 & 66.90 & 62.33 & 78.76/80.36 & 57.53/2.06 & 59.12 \\ 
    RL-KD \citep{RN7370} & 59.78/58.99 & 58.33 & 66.03 & 69.07 & 77.93 & 65.78 & 76.87/82.74 & 69.24/22.21 & 65.26 \\
    Uncertainty \citep{RN7419} & 58.52/57.67 & 59.33 & 64.10 & 70.16 & 77.55 & 65.78 & 80.85/83.33 & 69.47/22.49 & 65.39 \\ \hline
    \multicolumn{10}{c}{\textbf{\textit{Teacher assistants}}: Teacher (GLM$_\mathrm{Large}$) $\Rightarrow$ Assistant (200M) $\Rightarrow$ Assistant (110M) $\Rightarrow$ Student (66M)} \\ \hline
    TAKD \citep{RN7475} & 25.50/24.69 & 60.33 & 66.03 & 55.11 & 66.39 & 57.94 & 76.28/76.79 & 55.90/1.50 & 54.52 \\ 
    DGKD \citep{RN7395} & 23.68/22.96 & 61.00 & 66.99 & 55.96 & 65.71 & 58.73 & 75.45/75.60 & 48.06/1.50 & 54.00 \\
    \hline
\end{tabular}
}
\caption{Results of 25 mainstream distillation methods implemented using GKD on the SuperGLUE validation set. Due to the alteration of the model structure by MobileBERT$^\dagger$, the parameters of the teacher and student models are 293M and 25M, respectively. $\Rightarrow$ denotes distillation process. The results for all methods were averaged over three random seeds.}
\label{tab:results}
\end{table*}

\paragraph{Larger-scale Model Distillation}
To verify the proposed teacher-student parallel strategy can support distillation of 100B-scale model on 8 NVIDIA A100 (40GB) GPUs, we present the memory and time consumption of different strategies for distilling models of varying scale in Table \ref{tab:memory-stat}. The results indicate that previous strategies encountered GPU memory overflow when distilling 6B-scale models, whereas our strategy is capable of supporting the distillation of 100B-scale models. The results in rows 9, 10, and 11 respectively demonstrate that GPU memory consumption can be reduced through splitting the model parameters, optimizer states, or storing the optimizer states in CPU memory. If not limited to 8 GPUs, our strategy has the potential to distill even larger models. Appendix \ref{sec:appendix:parallel} further examines the trade-off between memory and time consumption.

\begin{table*}
\centering
\resizebox{\linewidth}{!}{
\begin{tabular}{c|ccccccccc} \hline
	\textbf{Strategy} & \textbf{Teacher$\Rightarrow$Student (scale)} & \textbf{MA (GB)} & \textbf{CA (GB)} & \textbf{Time (ms)} & \textbf{Mem (GB)} & \textbf{MP} & \textbf{DP} & \textbf{ZeRO} & \textbf{Offload} \\ \hline
    \multirow{4}{*}{Previous} & 110M$\Rightarrow$22M & 0.99 & 1.27 & 10.40 & 56.96 & 1 & 8 &  &  \\ 
     & 110M$\Rightarrow$66M & 1.73 & 2.02 & 10.82 & 57.60 & 1 & 8 &  &  \\ 
     & 340M$\Rightarrow$66M & 3.11 & 3.58 & 16.41 & 63.46 & 1 & 8 &  &  \\ 
     & 5B$\Rightarrow$1B & 32.44 & 36.57 & 53.34 & 61.58 & 1 & 8 &  &  \\ 
     & 6B$\Rightarrow$1.2B & \multicolumn{4}{c}{GPU memory overflow} & 1 & 8 &  &  \\ \hline
    \multirow{11}{*}{Ours} & 6B$\Rightarrow$1.2B & 18.91 & 21.40 & 85.61 & 57.28 & 2 & 4 &  &  \\ 
     & 7.5B$\Rightarrow$1.5B & 24.22 & 27.36 & 87.08 & 60.44 & 2 & 4 &  &  \\ 
     & 10B$\Rightarrow$2B & 30.91 & 34.54 & 105.40 & 62.33 & 2 & 4 &  &  \\ 
     & 10B$\Rightarrow$2B & 18.45 & 22.56 & 119.72 & 68.68 & 2 & 4 & $\checkmark$ &  \\ 
     & 10B$\Rightarrow$2B & 15.83 & 22.55 & 387.19 & 106.35 & 2 & 4 & $\checkmark$ & $\checkmark$ \\ 
     & 25B$\Rightarrow$5B & 20.41 & 23.51 & 379.38 & 63.07 & 8 & 1 &  &  \\ 
     & 50B$\Rightarrow$10B & 17.93 & 20.94 & 4570.54 & 230.27 & 8 & 1 & $\checkmark$ & $\checkmark$ \\ 
    & 65B$\Rightarrow$13B & 22.48 & 26.10 & 6412.11 & 293.11 & 8 & 1 & $\checkmark$ & $\checkmark$ \\ 
    & 90B$\Rightarrow$18B & 30.56 & 35.27 & 7193.26 & 373.81 & 8 & 1 & $\checkmark$ & $\checkmark$ \\ 
     & 100B$\Rightarrow$20B & 33.62 & 36.88 & 9081.97 & 410.83 & 8 & 1 & $\checkmark$ & $\checkmark$ \\ 
     & 110B$\Rightarrow$22B & \multicolumn{4}{c}{GPU memory overflow} & 8 & 1 & $\checkmark$ & $\checkmark$ \\ 
    \hline
\end{tabular}
}
\caption{The consumption of memory and time during the pre-training stage of TinyBERT when distilling teacher models of different scales on 8 NVIDIA A100 (40GB) GPUs is presented. The micro batch and gradient accumulation steps are set to 1. Where MA denotes the maximum memory allocated on the GPU, CA denotes the maximum cached memory on the GPU, Time denotes the time required to train each sample, Mem denotes the size of occupied CPU memory, MP denotes the number of model parallelism, DP denotes the number of data parallelism, ZeRO denotes whether the optimizer states are partitioned across different GPUs, and Offload denotes whether the optimizer states are stored in CPU memory.}
\label{tab:memory-stat}
\end{table*}

\subsection{Further Exploration}
In addition to compatibility with various methods, GKD also allows for effortless combination of different methods. In Appendix \ref{sec:appendix:combination}, we have discovered a method that achieves SOTA results by combining the advantages of different distillation methods. Appendix \ref{sec:appendix:interpretability} presents a tool that analyzes the correlation between feature distance and task performance through GKD, enhancing the interpretability of the distillation process.

\section{Conclusions}
In this paper, we propose a general knowledge distillation framework, GKD, for deploying knowledge distillation systems targeting large-scale PLMs. GKD satisfies the demands of real-world applications by employing a parallel strategy and adaptive architecture, allowing for the distillation of ultra-large scale PLMs (over 10B) and the switch of various advanced distillation methods. In the future, we plan to launch our knowledge distillation system for facilitating the mass production and deployment of student models.

\section*{Acknowledgements}
This work is supported by Technology and Innovation Major Project of the Ministry of Science and Technology of China under Grant 2020AAA0108400 and 2020AAA0108402, the Natural Science Foundation of China under Grant No. 61836013, the Major Program of the National Social Science Foundation of China under Grant No. 18ZDA032, and funds from CCF-Zhipu.AI and Beijing Academy of Artificial Intelligence (BAAI). The GPUs used are sponsored by Zhipu.AI.

\bibliography{acl2023}
\bibliographystyle{acl_natbib}

\appendix

\section{Further Exploration}
In this section, we further explore the capabilities of GKD in combining different distillation methods and enhancing the interpretability of the distillation process.

\subsection{Method Combination}
\label{sec:appendix:combination}
Thanks to the dynamic hook mechanism, GKD is capable of combining methods by integrating hooks from different methods. As shown in Table \ref{tab:appendix:baselines}, we demonstrate results from several dozen combinations of different model features. To conserve computational power, we set the batch size to 32 during pre-training and set the sizes of the teacher and student models to 110M and 22M, respectively. In the task-specific stage, the batch size and learning rate were fixed at 16 and 1e-5, respectively, without the use of grid search and seed averaging. Based on the results in Table \ref{tab:appendix:baselines}, the following conclusions can be drawn.

\begin{table*}[!t]
\centering
\resizebox{\linewidth}{!}{
\begin{tabular}{l|ccccccc|ccccccc|c} \hline
    \multirow{2}{*}{\textbf{Methods}} & \multicolumn{7}{|c|}{\textbf{Pre-training stage}} & \multicolumn{7}{|c|}{\textbf{Task-specific stage}} & \multirow{2}{*}{\textbf{SG}}\\ \cline{2-15}
     & \textbf{Emb} & \textbf{Att} & \textbf{Q/K} & \textbf{V} & \textbf{HS} & \textbf{Soft} & \textbf{Hard} & \textbf{Emb} & \textbf{Att} &  \textbf{Q/K} & \textbf{V} & \textbf{HS} & \textbf{Soft} & \textbf{Hard} \\ \hline
    KD & \multicolumn{7}{|c|}{Random initialization parameters} &  &  &  &  &  & CE & CE & 49.48 \\ 
    & \multicolumn{7}{|c|}{Truncate fine-tuned teacher parameters} &  &  &  &  &  & CE & CE & 51.62 \\ 
     &  &  &  &  &  & CE & CE &  &  &  &  &  & CE & CE & 59.68 \\ 
     &  &  &  &  &  & CE & CE &  &  &  &  &  &  & CE & 60.24 \\ 
     &  &  &  &  &  & KL & CE &  &  &  &  &  &  & CE & 60.62 \\ 
     &  &  &  &  &  & KL &  &  &  &  &  &  &  & CE & 63.16 \\ 
     &  &  &  &  &  & MSE &  &  &  &  &  &  &  & CE & 63.46 \\ 
    RAIL-KD & \multicolumn{7}{|c|}{Truncate fine-tuned teacher parameters} &  &  &  &  & MSE$_{-f}$ & CE & CE & 51.63 \\ 
    MiniLM &  & KL$_{f}$ &  & KL$_{f}$ &  &  &  &  &  &  &  &  &  & CE & 60.65 \\ 
    MiniLMv2 &  &  & KL$_{f}$ & KL$_{f}$ &  &  &  &  &  &  &  &  &  & CE & 60.47 \\ 
     &  &  & KL$_{f}$ & KL$_{f}$ &  & KL &  &  &  & KL$_{f}$ & KL$_{f}$ &  &  & CE & 65.41 \\ 
     &  &  & KL$_{f}$ & KL$_{f}$ &  & KL &  &  &  &  &  &  &  & CE & 64.64 \\ 
    MGSKD & MSE & MSE &  &  & MSE &  &  & MSE/HL &  &  &  & MSE/HL & KL &  & 59.65 \\ 
    TinyBERT & MSE & MSE &  &  & MSE &  &  & MSE & MSE &  &  & MSE & CE &  & 65.81 \\ 
     & MSE & MSE &  &  & MSE & KL &  & MSE & MSE &  &  & MSE & CE &  & 66.19 \\ 
     & MSE & MSE &  &  & MSE & KL &  &  &  &  &  &  & CE &  & 62.75 \\ 
     & MSE & MSE &  &  & MSE &  &  &  &  &  &  &  & CE &  & 63.52 \\ 
     & MSE &  &  &  & MSE & KL &  & MSE &  &  &  & MSE & CE &  & 66.51 \\ 
    Mix5 & MSE & MSE+KL$_{f}$ & KL$_{f}$ & KL$_{f}$ & MSE+Cos & KL & CE & MSE & MSE+KL$_{f}$ & KL$_{f}$ & KL$_{f}$ & MSE & CE & CE & 65.63 \\ 
     & MSE & MSE+KL$_{f}$ & KL$_{f}$ & KL$_{f}$ & MSE+Cos & KL & CE &  &  &  &  &  &  & CE & 62.18 \\ 
     & MSE & MSE & KL$_{f}$ & KL$_{f}$ & MSE+Cos & KL & CE & MSE & MSE & KL$_{f}$ & KL$_{f}$ & MSE & CE & CE & 66.58 \\ 
     & MSE & MSE & KL$_{f}$ & KL$_{f}$ & MSE+Cos & KL & CE &  &  &  &  &  &  & CE & 63.06 \\ 
     & MSE & MSE+KL$_{f}$ &  & KL$_{f}$ & MSE+Cos & KL & CE & MSE & MSE+KL$_{f}$ &  & KL$_{f}$ & MSE & CE & CE & 66.25 \\ 
     & MSE & MSE+KL$_{f}$ &  & KL$_{f}$ & MSE+Cos & KL & CE &  &  &  &  &  &  & CE & 62.86 \\ 
     & MSE & MSE+KL$_{f}$ & KL$_{f}$ & KL$_{f}$ & MSE &  & CE & MSE & MSE+KL$_{f}$ & KL$_{f}$ & KL$_{f}$ & MSE & CE & CE & 66.54 \\ 
     & MSE & MSE+KL$_{f}$ & KL$_{f}$ & KL$_{f}$ & MSE &  & CE &  &  &  &  &  &  & CE & 64.64 \\ 
     &  & KL$_{f}$ & KL$_{f}$ & KL$_{f}$ &  & KL & CE &  & KL$_{f}$ & KL$_{f}$ & KL$_{f}$ &  & CE & CE & 64.68 \\ 
     &  & KL$_{f}$ & KL$_{f}$ & KL$_{f}$ &  & KL & CE &  &  &  &  &  &  & CE & 60.49 \\ 
    BestC & MSE &  & KL$_{f}$ & KL$_{f}$ & MSE & KL &  & MSE &  & KL$_{f}$ & KL$_{f}$ & MSE & CE &  & \bf 67.05 \\ 
     & MSE &  & KL$_{f}$ & KL$_{f}$ & MSE & KL &  &  &  &  &  &  &  & CE & 62.71 \\ 
     & MSE &  & KL$_{f}$ & KL$_{f}$ & MSE$_{f}$ & KL &  & MSE &  & KL$_{f}$ & KL$_{f}$ & MSE$_{f}$ & CE &  & 65.73 \\ 
     & MSE &  & KL$_{f}$ & KL$_{f}$ & MSE$_{f}$ & KL &  &  &  &  &  &  &  & CE & 62.53 \\ 
     & MSE &  &  &  & MSE$_{f}$ & KL &  & MSE &  &  &  & MSE$_{f}$ & CE &  & 66.17 \\ 
     & MSE &  &  &  & MSE$_{f}$ & KL &  &  &  &  &  &  &  & CE & 62.58 \\ 
     &  &  &  &  & MSE$_{f}$ & KL &  &  &  &  &  & MSE$_{f}$ & CE &  & 65.79 \\ 
     &  &  &  &  & MSE$_{f}$ & KL &  &  &  &  &  &  &  & CE & 63.69 \\ 
     &  &  & KL$_{f}$ & KL$_{f}$ & MSE$_{f}$ & KL &  &  &  & KL$_{f}$ & KL$_{f}$ & MSE$_{f}$ & CE &  & 66.01 \\ 
     &  &  & KL$_{f}$ & KL$_{f}$ & MSE$_{f}$ & KL &  &  &  &  &  &  &  & CE & 63.87 \\ 
     & MSE &  & KL$_{f}$ & KL$_{f}$ &  & KL &  & MSE &  & KL$_{f}$ & KL$_{f}$ &  & CE &  & 66.53 \\ 
     & MSE &  & KL$_{f}$ & KL$_{f}$ &  & KL &  &  &  &  &  &  &  & CE & 63.26 \\ 
     & MSE &  & KL$_{f}$ & KL$_{f}$ & MSE$_{f2}$ & KL &  & MSE &  & KL$_{f}$ & KL$_{f}$ & MSE$_{f2}$ & CE &  & 65.55 \\ 
     & MSE &  & KL$_{f}$ & KL$_{f}$ & MSE$_{f2}$ & KL &  &  &  &  &  &  &  & CE & 62.56 \\ 
     & MSE &  & KL$_{f}$ & KL$_{f}$ & MSE$_{-f}$ & KL &  & MSE &  & KL$_{f}$ & KL$_{f}$ & MSE$_{-f}$ & CE &  & 66.22 \\ 
     & MSE &  & KL$_{f}$ & KL$_{f}$ & MSE$_{-f}$ & KL &  &  &  &  &  &  &  & CE & 63.26 \\ 
     &  &  & KL$_{f}$ & KL$_{f}$ & MSE & KL &  &  &  & KL$_{f}$ & KL$_{f}$ & MSE & CE &  & 66.78 \\ 
     &  &  & KL$_{f}$ & KL$_{f}$ & MSE & KL &  &  &  &  &  &  &  & CE & 63.12 \\ 
    \hline
\end{tabular}
}
\caption{Results of combining various features of models using GKD on the SuperGLUE validation set. Emb, Att, Q/K, V, HS, Soft, Hard, and SG denote the output of the embedding layer, attention scores, query/key matrix, value matrix, hidden state, soft labels, hard labels, and the average score on the SuperGLUE benchmark, respectively. MSE, KL, CE, Cos, and HL respectively denote the distance functions between the teacher and student features as mean squared error, Kullback-Leibler divergence, cross-entropy, cosine distance, and Huber loss. MSE$_{f}$, MSE$_{-f}$, and MSE$_{f2}$ respectively indicate the calculation of MSE for the last layer, before the last layer, and the second-to-last layer's hidden state. MSE+KL$_{f}$ represents the sum of MSE and KL calculated for the last layer's attention scores. The Mix5 method can be understood as a combination of the KD \citep{RN4609}, TinyBERT \citep{RN7482}, MiniLM \citep{RN7471}, MiniLMv2 \citep{RN7385}, and DistilBERT \citep{RN6418} methods.}
\label{tab:appendix:baselines}
\end{table*}

(1) We discovered the method BestC which achieves the SOTA, outperforming TinyBERT by 1.24\% on average in SuperGLUE. BestC combines the features of TinyBERT, MiniLMv2, and soft labels. (2) The method that performs distillation in the pre-training stage (row 5) outperforms those using randomly initialized parameters (row 3) or truncated fine-tuned teacher parameters (row 4) in the pre-training stage. (3) The methods using soft labels in the pre-training stage (rows 14 and 17) outperform those not using soft labels (rows 12 and 16). (4) Starting from row 21, we compare the results of various combinations distilled in the task-specific stage and not distilled (only trained on hard labels). We find that distillation in the task-specific stage greatly improves the performance of the task.

\begin{figure*}[t]
  \centering
  \includegraphics[width=\linewidth,scale=1]{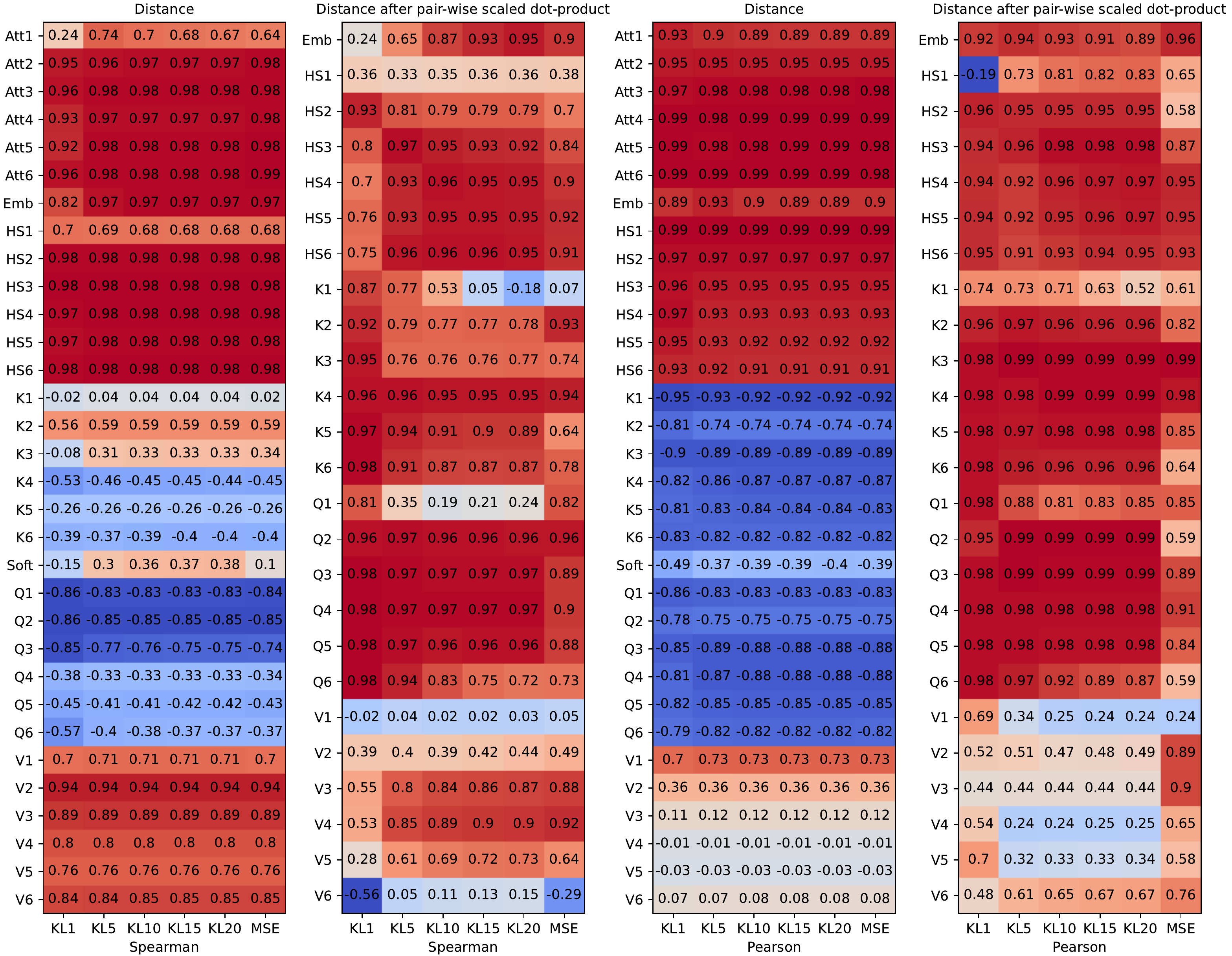}
  \caption{Spearman and Pearson correlation coefficient of pre-training loss with the distance between teacher and student features of TinyBERT. This records the training process of TinyBERT in Table \ref{tab:results}, where the sizes of the teacher and student models are 110M and 66M respectively. The distance after pair-wise scaled dot-product is calculated by first computing features $\mathbf H \leftarrow \frac{\mathbf H\mathbf H^T}{\sqrt{dimensionality}}$. Att1, HS1, Q1, K1, and V1 denote the attention scores, hidden state, query matrix, key matrix, and value matrix of the first layer transformer, respectively. Soft and Emb denote the soft labels and output of the embedding-layer respectively. KL1, KL5, KL10, KL15, and KL20 denote the KL divergence with temperatures of 1, 5, 10, 15, and 20, respectively. \label{fig:appendix:tinybert-loss}}
\end{figure*}

\begin{figure*}[t]
  \centering
  \includegraphics[width=\linewidth,scale=1]{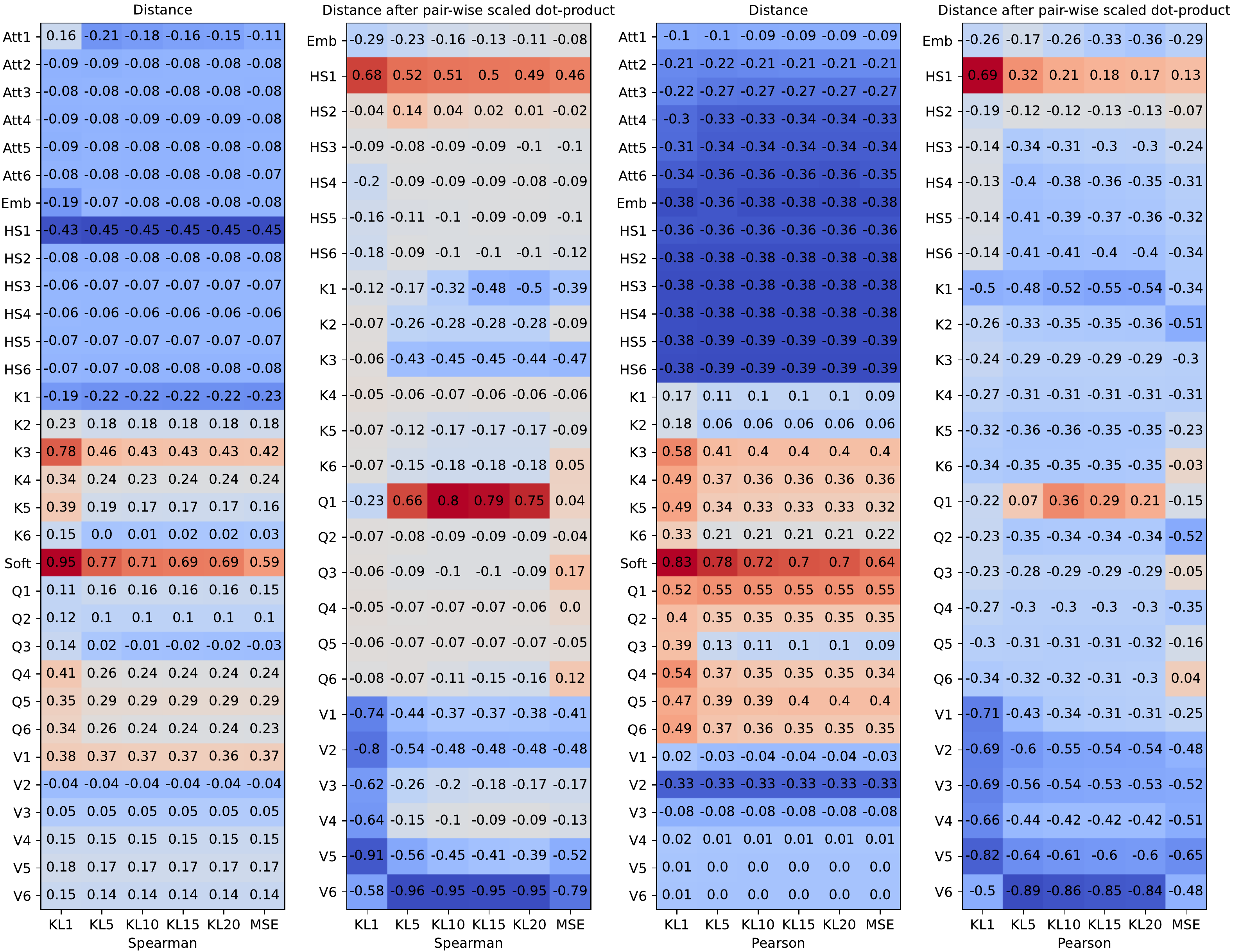}
  \caption{Spearman and Pearson correlation coefficient of task performance (perplexity of the language model on the validation set) of pre-training stage with the distance between teacher and student features of TinyBERT. This records the training process of TinyBERT in Table \ref{tab:results}, where the sizes of the teacher and student models are 110M and 66M respectively. The distance after pair-wise scaled dot-product is calculated by first computing features $\mathbf H \leftarrow \frac{\mathbf H\mathbf H^T}{\sqrt{dimensionality}}$. Att1, HS1, Q1, K1, and V1 denote the attention scores, hidden state, query matrix, key matrix, and value matrix of the first layer transformer, respectively. Soft and Emb denote the soft labels and output of the embedding-layer respectively. KL1, KL5, KL10, KL15, and KL20 denote the KL divergence with temperatures of 1, 5, 10, 15, and 20, respectively. \label{fig:appendix:tinybert-ppl}}
\end{figure*}

\subsection{Enhanced Interpretability}
\label{sec:appendix:interpretability}
Thanks to the adaptive architecture, GKD can record all model features through extraction hooks and save the distance of teacher and student features in the auxiliary model for later analysis of the correlation between feature distance and task performance in the distillation process. As an example of TinyBERT's pre-training stage distillation, we present the Spearman and Pearson correlation coefficients between the feature distance and training loss, and between the feature distance and task performance, respectively, in Figures \ref{fig:appendix:tinybert-loss} and \ref{fig:appendix:tinybert-ppl}. The following conclusions can be drawn.

(1) The results shown in Figure \ref{fig:appendix:tinybert-loss} indicate that while TinyBERT trains its embedding layer, attention scores, and hidden state, many other features (e.g., the value matrix and features obtained after pair-wise scaled dot-product) also decrease in distance between teacher and student as the training loss decreases. This suggests that we may be able to find a way to automatically have a large number of student features approach the teacher without having to distill all features, thus reducing the cost of distillation. (2) The results shown in Figure \ref{fig:appendix:tinybert-ppl} indicate that the distillation in the pre-training stage of TinyBERT actually leads to a decrease in pre-training task performance. This suggests that the performance of pre-training tasks is not necessarily positively correlated with the performance of downstream tasks. It is noteworthy that there are a small number of features (e.g., soft labels) whose distance is related to task performance. Our hypothesis is that distilling features that are related to task performance may further improve task performance, and the third conclusion in Appendix \ref{sec:appendix:combination} supports this hypothesis.

\section{Additional Analysis}
In this section, we further verify the reliability of GKD from the perspective of loss function value, and analyze the balance of memory and time consumption in the teacher-student parallel strategy.

\subsection{Are the Loss Values of GKD Normal?}
\label{sec:appendix:loss}
\begin{figure*}
  \centering
  \includegraphics[width=\linewidth,scale=1]{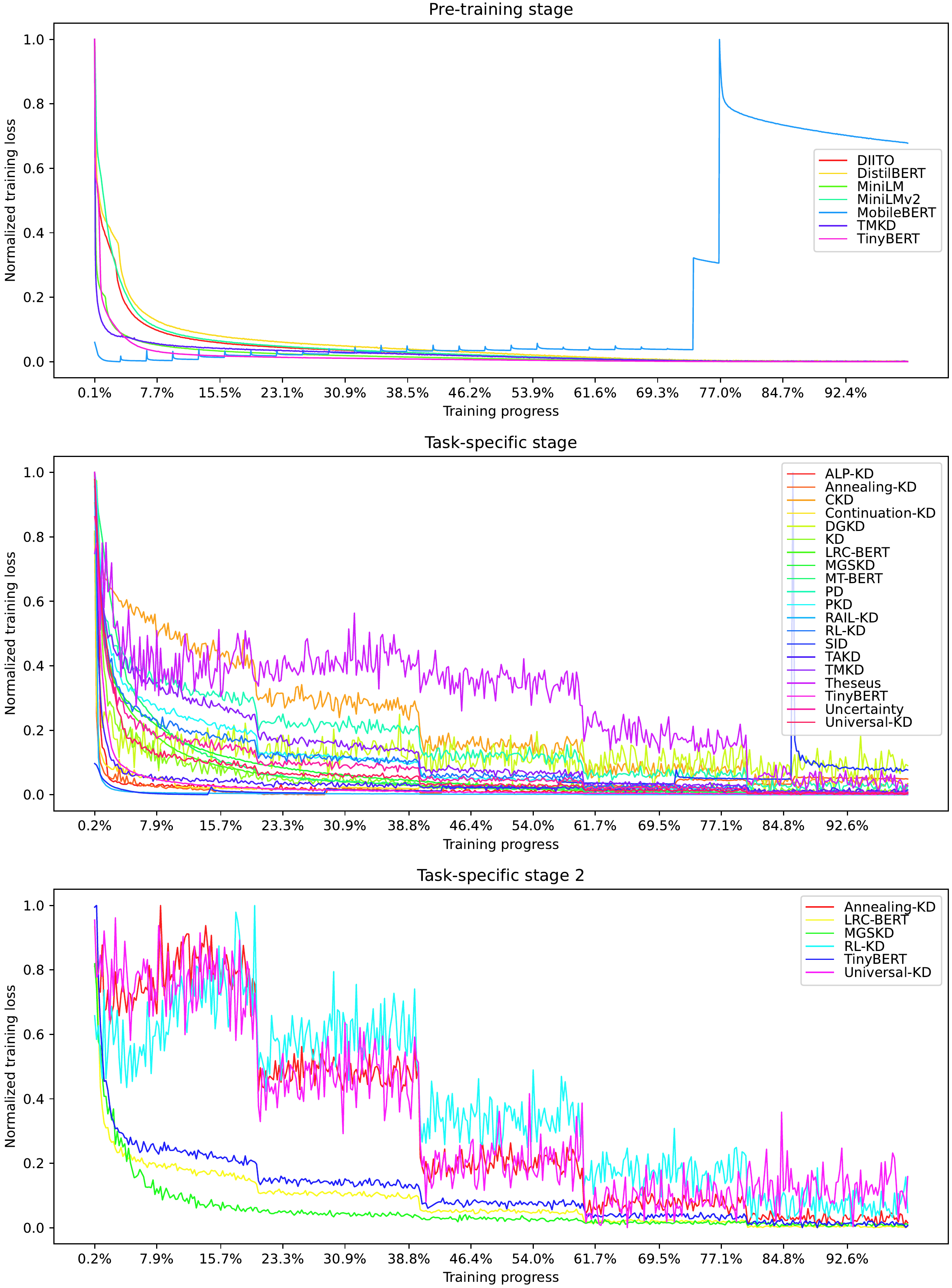}
  \caption{Loss function values of 25 methods across different distillation stages. The loss values are normalized due to the varying range of values across different methods. Some methods are distilled at most 3 times, including a pre-training stage and two task-specific stages (ReCoRD dataset). TAKD and DGKD based on teacher-assistant strategy showcase the final distillation process. \label{fig:appendix:method-loss}}
\end{figure*}

In order to further verify the reliability of GKD, we present the loss function values of each method at various distillation stages in Figure \ref{fig:appendix:method-loss}. The downward trend of all the loss values is consistent with our expectations, with two noteworthy observations: (1) MobileBERT and SID tend to gradually increase the number of distilled layers during training, hence the loss values exhibit an up-and-down trend. (2) The ReCoRD dataset, shown in task-specific stages, was trained for 5 epochs, therefore some methods may show loss changes in stair-step fashion, such as Annealing-KD and Universal-KD.

\subsection{Trade-off between Memory and Time Consumption}
\label{sec:appendix:parallel}

\begin{table*}
\centering
\resizebox{0.9\linewidth}{!}{
\begin{tabular}{c|cccccccc} \hline
    \textbf{Teacher$\Rightarrow$Student (scale)} & \textbf{MA (GB)} & \textbf{CA (GB)} & \textbf{Time (ms)} & \textbf{Mem (GB)} & \textbf{MP} & \textbf{DP} & \textbf{ZeRO} & \textbf{Offload} \\ \hline
    \multirow{20}{*}{5B$\Rightarrow$1B} & 17.65 & 33.11 & 169.01 & 95.15 & 1 & 8 & $\checkmark^\dagger$ & $\checkmark^\dagger$ \\
    & 9.07 & 14.97 & 262.49 & 86.56 & 2 & 4 & $\checkmark^\dagger$ & $\checkmark^\dagger$ \\
    & 4.87 & 6.09 & 430.61 & 86.20 & 4 & 2 & $\checkmark^\dagger$ & $\checkmark^\dagger$ \\
    & \bf 2.72 & 3.71 & 884.05 & 82.61 & 8 & 1 & $\checkmark^\dagger$ & $\checkmark^\dagger$ \\
    & 17.65 & 30.77 & 175.08 & 95.13 & 1 & 8 & $\checkmark$ & $\checkmark$ \\
    & 9.06 & 13.74 & 252.99 & 86.74 & 2 & 4 & $\checkmark$ & $\checkmark$ \\
    & 4.87 & 5.79 & 437.50 & 86.01 & 4 & 2 & $\checkmark$ & $\checkmark$ \\
    & \bf 2.72 & \bf 3.43 & 831.22 & 83.24 & 8 & 1 & $\checkmark$ &  $\checkmark$\\
    & 18.92 & 31.04 & 61.43 & 70.50 & 1 & 8 & $\checkmark^\dagger$ &  \\
    & 10.44 & 14.24 & 78.97 & 62.18 & 2 & 4 & $\checkmark^\dagger$ &  \\
    & 6.31 & 7.42 & 129.91 & 61.59 & 4 & 2 & $\checkmark^\dagger$ &  \\
    & 4.23 & 5.10 & 260.96 & \bf 58.72 & 8 & 1 & $\checkmark^\dagger$ &  \\
    & 18.92 & 28.81 & 60.25 & 70.52 & 1 & 8 & $\checkmark$ &  \\
    & 10.43 & 13.73 & 80.64 & 62.38 & 2 & 4 & $\checkmark$ &  \\
    & 6.31 & 7.32 & 129.40 & 62.27 & 4 & 2 & $\checkmark$ &  \\
    & 4.23 & 5.04 & 243.80 & 58.76 & 8 & 1 & $\checkmark$ &  \\
    & 32.44 & 36.57 & \bf 53.34 & 61.58 & 1 & 8 &  &  \\
    & 16.34 & 18.50 & 68.17 & 62.18 & 2 & 4 &  &  \\
    & 8.31 & 9.41 & 121.26 & 62.16 & 4 & 2 &  &  \\
    & 4.27 & 5.04 & 231.95 & 58.83 & 8 & 1 &  &  \\ \hline
    \multirow{20}{*}{10B$\Rightarrow$2B} & 30.73 & 36.89 & 226.51 & 104.46 & 1 & 8 & $\checkmark^\dagger$ & $\checkmark^\dagger$ \\
    & 15.84 & 24.19 & 378.93 & 106.31 & 2 & 4 & $\checkmark^\dagger$ & $\checkmark^\dagger$ \\
    & 8.41 & 10.08 & 664.45 & 95.51 & 4 & 2 & $\checkmark^\dagger$ & $\checkmark^\dagger$ \\
    & \bf 4.70 & 6.00 & 1210.07 & 98.12 & 8 & 1 & $\checkmark^\dagger$ & $\checkmark^\dagger$ \\
    & 30.73 & 36.89 & 222.50 & 104.45 & 1 & 8 & $\checkmark$ & $\checkmark$ \\
    & 15.83 & 22.55 & 387.19 & 106.35 & 2 & 4 & $\checkmark$ & $\checkmark$ \\
    & 8.41 & 9.72 & 693.03 & 95.50 & 4 & 2 & $\checkmark$ & $\checkmark$ \\
    & \bf 4.70 & \bf 5.81 & 1224.11 & 98.09 & 8 & 1 & $\checkmark$ & $\checkmark$ \\
    & 33.23 & 36.91 & \bf 85.52 & 66.17 & 1 & 8 & $\checkmark^\dagger$ &  \\
    & 18.46 & 23.13 & 119.30 & 68.61 & 2 & 4 & $\checkmark^\dagger$ &  \\
    & 11.11 & 12.91 & 186.53 & \bf 57.42 & 4 & 2 & $\checkmark^\dagger$ &  \\
    & 7.53 & 8.87 & 310.56 & 59.88 & 8 & 1 & $\checkmark^\dagger$ &  \\
    & 33.23 & 36.90 & 88.21 & 66.13 & 1 & 8 & $\checkmark$ &  \\
    & 18.45 & 22.56 & 119.72 & 68.68 & 2 & 4 & $\checkmark$ &  \\
    & 11.11 & 12.78 & 198.84 & 57.45 & 4 & 2 & $\checkmark$ &  \\
    & 7.53 & 9.01 & 329.12 & 59.83 & 8 & 1 & $\checkmark$ &  \\
    & \multicolumn{4}{c}{GPU memory overflow} & 1 & 8 &  &  \\
    & 30.91 & 34.54 & 105.40 & 62.33 & 2 & 4 &  &  \\
    & 15.64 & 17.62 & 174.30 & 57.79 & 4 & 2 &  &  \\
    & 8.00 & 8.98 & 311.44 & 59.73 & 8 & 1 &  &  \\
    \hline
\end{tabular}
}
\caption{The consumption of memory and time during the pre-training stage of TinyBERT when distilling teacher models of different scales on 8 NVIDIA A100 (40GB) GPUs is presented. The micro batch and gradient accumulation steps are set to 1. Where MA denotes the maximum memory allocated on the GPU, CA denotes the maximum cached memory on the GPU, Time denotes the time required to train each sample, Mem denotes the size of occupied CPU memory, MP denotes the number of model parallelism, DP denotes the number of data parallelism, ZeRO denotes whether the optimizer states are partitioned across different GPUs, and Offload denotes whether the optimizer states are stored in CPU memory. In addition to the optimizer states, the model gradients can also be partitioned across different GPUs or stored in CPU memory. The dagger symbol ($^\dagger$) represents optimization of both the optimizer states and the model gradients simultaneously.}
\label{tab:appendix:memory-stat}
\end{table*}

\begin{table*}
\centering
\resizebox{0.77\linewidth}{!}{
\begin{tabular}{l|cccccccc} \hline
	\textbf{Hyperparameters} & \textbf{ReCoRD} & \textbf{COPA} & \textbf{WSC} & \textbf{RTE} & \textbf{BoolQ} & \textbf{WiC} & \textbf{CB} & \textbf{MultiRC} \\ \hline
    Sequence length & 512 & 256 & 128 & 256 & 256 & 256 & 256 & 512 \\ 
    Epochs & 5 & 50 & 50 & 50 & 20 & 30 & 50 & 15 \\ 
    Dropout & \multicolumn{8}{|c}{0.1} \\ 
    Attention Dropout & \multicolumn{8}{|c}{0.1} \\ 
    Warmup Ration & \multicolumn{8}{|c}{0.1} \\ 
    Weight Decay & \multicolumn{8}{|c}{0.1} \\ 
    Learning Rate Decay & \multicolumn{8}{|c}{Linear} \\ 
    Adam $\epsilon$ & \multicolumn{8}{|c}{1E-8} \\ 
    Adam $\beta_1$ & \multicolumn{8}{|c}{0.9} \\ 
    Adam $\beta_2$ & \multicolumn{8}{|c}{0.999} \\ 
    Gradient Clipping & \multicolumn{8}{|c}{0.1} \\ 
    \hline
\end{tabular}
}
\caption{Other hyperparameters for the task-specific stage on the 8 datasets of the SuperGLUE benchmark.}
\label{tab:appendix:other-hyper}
\end{table*}

In order to speed up the training process while ensuring that the distillation process is not limited by GPU memory, we conducted a full combination of all optimization options to find the best balance between memory and time. Table \ref{tab:appendix:memory-stat} showcases the resource usage of 5B-scale and 10B-scale teacher models under different MP, DP, ZeRO, and Offload options during distillation. The results of the testing lead us to the following recommendations: In cases of insufficient GPU memory, ZeRO should be considered first for partitioning the optimizer states and model gradients, followed by increasing the number of model parallelism, and lastly, using ZeRO-Offload to store the optimizer states and model gradients in CPU memory.

\section{Implementation Details}
\label{sec:appendix:details}
In this section, we provide further details regarding the hyperparameters and models to facilitate replication by developers.

\subsection{Hyperparameters}
\label{sec:appendix:hyper}
The batch size, number of iterations, and peak learning rate for the pre-training stage were set to 64, 150000, and 4e-4, respectively. The task-specific hyperparameters for specific methods were set to the optimal values from their corresponding papers, while other hyperparameters (see Table \ref{tab:appendix:other-hyper}) were kept consistent with the fine-tuning teacher. For single-teacher methods in the task-specific stage, grid search was used to optimize hyperparameters, including learning rate \{5e-6,1e-5,2e-5\} and batch size \{16,32\}. Table~\ref{tab:appendix:hyper} presents the learning rate and batch size for each method on each dataset in the SuperGLUE benchmark. The results for all methods were averaged over three random seeds.

\begin{table*}
\centering
\resizebox{\linewidth}{!}{
\begin{tabular}{l|cccccccc} \hline
    \multirow{2}{*}{\textbf{Methods}} & \textbf{ReCoRD} & \textbf{COPA} & \textbf{WSC} & \textbf{RTE} & \textbf{BoolQ} & \textbf{WiC} & \textbf{CB} & \textbf{MultiRC} \\ \cline{2-9}
     & \textbf{bs/lr}& \textbf{bs/lr}& \textbf{bs/lr}& \textbf{bs/lr}& \textbf{bs/lr}& \textbf{bs/lr}& \textbf{bs/lr}& \textbf{bs/lr} \\ \hline
    GLM$_\mathrm{Base}$ (teacher, 110M) & \multicolumn{8}{|c}{\multirow{2}{*}{bs (batch size) = 16, lr (learning rate) = 1E-5}} \\
    GLM$_\mathrm{Large}$ (teacher, 340M) &  &  &  &  &  &  &  &   \\ \hline
    \multicolumn{9}{c}{\textbf{\textit{Single-teacher}}: Teacher (GLM$_\mathrm{Base}$) $\Rightarrow$ Student (66M)} \\ \hline
    KD \citep{RN4609} & 16/5E-06 & 16/2E-05 & 16/1E-05 & 16/2E-05 & 16/2E-05 & 16/5E-06 & 16/2E-05 & 16/5E-06 \\ 
    PD \citep{RN6581} & 16/1E-05 & 32/5E-06 & 16/2E-05 & 16/1E-05 & 32/1E-05 & 16/5E-06 & 16/2E-05 & 16/5E-06 \\ 
    PKD \citep{RN7494} & 32/2E-05 & 32/2E-05 & 16/2E-05 & 32/5E-06 & 16/1E-05 & 16/5E-06 & 16/2E-05 & 32/2E-05 \\ 
    DistilBERT \citep{RN6418} & 16/1E-05 & 16/2E-05 & 16/1E-05 & 16/5E-06 & 32/2E-05 & 32/2E-05 & 32/2E-05 & 16/1E-05 \\ 
    Theseus \citep{RN7113} & 32/2E-05 & 16/1E-05 & 16/1E-05 & 32/1E-05 & 16/1E-05 & 32/1E-05 & 16/2E-05 & 32/5E-06 \\ 
    TinyBERT \citep{RN7482} & 32/1E-05 & 16/5E-06 & 32/5E-06 & 16/2E-05 & 16/1E-05 & 16/5E-06 & 16/1E-05 & 16/1E-05 \\ 
    MobileBERT \citep{RN7111} & 16/1E-05 & 16/1E-05 & 32/2E-05 & 32/2E-05 & 32/2E-05 & 32/1E-05 & 32/2E-05 & 16/5E-06 \\
    SID \citep{RN7487} & 16/2E-05 & 32/5E-06 & 16/5E-06 & 16/2E-05 & 16/2E-05 & 16/2E-05 & 16/1E-05 & 16/2E-05 \\ 
    MiniLM \citep{RN7471} & 16/2E-05 & 32/1E-05 & 32/2E-05 & 32/1E-05 & 16/1E-05 & 16/1E-05 & 32/1E-05 & 32/2E-05 \\ 
    MiniLMv2 \citep{RN7385} & 16/1E-05 & 16/1E-05 & 16/5E-06 & 32/2E-05 & 16/2E-05 & 32/2E-05 & 16/1E-05 & 16/1E-05 \\ 
    ALP-KD \citep{RN7406} & 16/2E-05 & 16/1E-05 & 16/2E-05 & 16/2E-05 & 16/2E-05 & 32/2E-05 & 16/2E-05 & 32/2E-05 \\ 
    LRC-BERT \citep{RN7437} & 16/2E-05 & 32/1E-05 & 16/2E-05 & 32/1E-05 & 16/2E-05 & 16/5E-06 & 16/2E-05 & 16/5E-06 \\ 
    Annealing-KD \citep{RN5665} & 16/2E-05 & 16/5E-06 & 16/2E-05 & 16/2E-05 & 16/2E-05 & 32/5E-06 & 16/1E-05 & 32/5E-06 \\ 
    CKD \citep{RN7407} & 32/2E-05 & 16/2E-05 & 16/5E-06 & 16/1E-05 & 16/2E-05 & 16/1E-05 & 16/1E-05 & 32/2E-05 \\ 
    Universal-KD \citep{RN7376} & 32/2E-05 & 32/5E-06 & 32/5E-06 & 32/1E-05 & 32/5E-06 & 16/5E-06 & 16/1E-05 & 16/1E-05 \\ 
    DIITO \citep{RN7325} & 16/5E-06 & 32/1E-05 & 16/2E-05 & 16/1E-05 & 16/2E-05 & 16/1E-05 & 16/1E-05 & 16/5E-06 \\ 
    Continuation-KD \citep{Continuation} & 16/2E-05 & 32/1E-05 & 16/1E-05 & 16/1E-05 & 16/2E-05 & 32/1E-05 & 16/1E-05 & 16/5E-06 \\ 
    RAIL-KD \citep{RN7350} & 16/1E-05 & 16/1E-05 & 16/2E-05 & 16/5E-06 & 32/2E-05 & 16/1E-05 & 32/1E-05 & 32/2E-05 \\ 
    MGSKD \citep{RN7341} & 16/5E-06 & 16/2E-05 & 32/2E-05 & 16/5E-06 & 16/5E-06 & 16/1E-05 & 32/2E-05 & 32/5E-06 \\  \hline
    \multicolumn{9}{c}{\textbf{\textit{Multi-teacher}}: Teachers (GLM$_\mathrm{Base}$ and GLM$_\mathrm{Large}$) $\Rightarrow$ Student (66M)} \\ \hline
    TMKD \citep{RN6109} & \multicolumn{8}{|c}{\multirow{4}{*}{same as GLM$_\mathrm{Base}$}} \\ 
    MT-BERT \citep{RN7379} &   &  &  &  &  &  &  &   \\ 
    RL-KD \citep{RN7370} &   &  &  &  &  &  &  &   \\
    Uncertainty \citep{RN7419} &   &  &  &  &  &  &  &   \\ \hline
    \multicolumn{9}{c}{\textbf{\textit{Teacher assistants}}: Teacher (GLM$_\mathrm{Large}$) $\Rightarrow$ Assistant (200M) $\Rightarrow$ Assistant (110M) $\Rightarrow$ Student (66M)} \\ \hline
    TAKD \citep{RN7475} & \multicolumn{8}{|c}{\multirow{2}{*}{same as KD}} \\ 
    DGKD \citep{RN7395} &   &  &  &  &  &  &  &   \\
    \hline
\end{tabular}
}
\caption{Hyperparameters for all methods in Table~\ref{tab:results} on the 8 datasets of the SuperGLUE benchmark.}
\label{tab:appendix:hyper}
\end{table*}

\subsection{Models}
\label{sec:appendix:models}
Table \ref{tab:appendix:models} shows the specific parameters of all the models utilized in this paper. The 110M, 340M, and 10B scale models are from GLM pre-trained models \footnote{https://github.com/THUDM/GLM}. The 293M-scale model with the MobileBERT structure (inverted-bottleneck structure) was obtained by us through a week of pre-training with 16 NVIDIA A100 (40GB) GPUs, and the 25M-scale model is also with the MobileBERT structure. When conducting pre-training tasks, the models with the MobileBERT structure require the expansion of the token dimension, thus the actual number of parameters is greater than the scale. The other sized teacher models were tested with randomly initialized parameters to assess resource consumption. All the distillation processes were conducted using half-precision floating-point (fp16) models.

\begin{table*}
\centering
\resizebox{0.75\linewidth}{!}{
\begin{tabular}{c|cccccc} \hline
	\textbf{Scale} & \textbf{\#Parameters} & \textbf{\#Dimensions} & \textbf{\#Layers} & \textbf{\#Heads} & \textbf{Max-seq} & \textbf{Vocabulary} \\ \hline
    22M & 22788864 & 384 & 6 & 12 & \multirow{6}{*}{512} & \multirow{6}{*}{30592} \\ 
    25M & 37371392 & 128 & 24 & 4 &  &  \\ 
    66M & 66811392 & 768 & 6 & 12 &  &  \\ 
    110M & 109338624 & 768 & 12 & 12 &  &  \\ 
    293M & 306174464 & 1024 & 24 & 4 &  &  \\ 
    340M & 334688256 & 1024 & 24 & 16 &   &  \\ \hline
    1B & 1022682240 & 1728 & 26 & \multirow{17}{*}{64} & \multirow{17}{*}{1024} & \multirow{17}{*}{50304} \\ 
    1.2B & 1173458944 & 1792 & 28 &  &  &  \\ 
    1.5B & 1521700224 & 1984 & 30 &  &  &  \\ 
    2B & 1920122880 & 2048 & 36 &  &  &  \\ 
    5B & 5030587776 & 3264 & 38 &  &  &  \\ 
    6B & 5915828736 & 3456 & 40 &  &  &  \\ 
    7.5B & 7385878656 & 3776 & 42 &  &  &  \\ 
    10B & 9880682496 & 4096 & 48 &  &  &  \\ 
    13B & 13170418176 & 4736 & 48 &  &  &  \\ 
    18B & 18125342976 & 5248 & 54 &  &  &  \\ 
    20B & 20175676160 & 5440 & 56 &  &  &  \\ 
    22B & 22104152064 & 5504 & 60 &  &  &  \\ 
    25B & 24660072448 & 5632 & 64 &  &  &  \\ 
    50B & 49577504000 & 8000 & 64 &  &  &  \\ 
    65B & 64813768448 & 9152 & 64 &  &  &  \\ 
    90B & 89957891328 & 10624 & 66 &  &  &  \\ 
    100B & 99465734144 & 11008 & 68 &  &  &  \\ 
    110B & 109620044032 & 11392 & 70 &  &  &  \\
    \hline
\end{tabular}
}
\caption{The scale details of all the models utilized in this paper.}
\label{tab:appendix:models}
\end{table*}

\end{document}